\documentclass{jedm} 

\usepackage{booktabs} 
\usepackage{ftnxtra}
\usepackage{fnpos}
\usepackage{lmodern}
\usepackage{amsmath}
\usepackage{amsfonts}
\usepackage{array}
\usepackage[caption=false,font=footnotesize]{subfig}
\usepackage{stfloats}
\usepackage{url}
\usepackage{multirow}
\usepackage[normalem]{ulem}
\usepackage{threeparttable}
\usepackage{hyperref}
\usepackage{booktabs}
\usepackage{graphicx}
\usepackage{subfig}
\usepackage{tikz}
\usepackage{amsmath}
\usepackage{mathtools}
\usepackage{amssymb}
\usepackage{color}
\usepackage{longtable}
\usepackage{array}
\usepackage{bm}

\newcommand{\nak}{NAK}
\newcommand{\mak}{MAK}
\newcommand{\cnak}{CNAK}
\newcommand{\cmak}{CMAK}
\newcommand{\ckrm}{CKRM}

\newcommand{\q}{\mbox{$\mathbf{q}$}}
\newcommand{\p}{\mbox{$\mathbf{p}$}}
\newcommand{\x}{\mbox{$\mathbf{x}$}}

\newcommand{\e}{\mbox{$\mathbf{e}$}}
\newcommand{\req}{\mbox{$\mathbf{r}$}}
\newcommand{\ks}{\mbox{$\mathbf{k}$}}

\newcommand{\term}{\mbox{$\mathcal{T}$}}

\newcommand{\Gmat}{\mbox{$\mathbf{G}$}}

\newcommand{\etal}{{\em et al.}}
\newcommand{\ul}{\underline}
\newcommand{\red}{\textcolor{red}}
\newcommand{\black}{\textcolor{black}}

\graphicspath{{./fig/}}
\DeclareGraphicsExtensions{.pdf}

\begin{document}
\title{Context-aware Non-linear and Neural Attentive Knowledge-based
  Models for Grade Prediction}
\date{}

\author{{\large Sara Morsy}\\
  Department of Computer Science\\
  and Engineering\\
  University of Minnesota\\
  morsy@cs.umn.edu \and
  {\large George Karypis}\\
  Department of Computer Science\\
  and Engineering\\
  University of Minnesota\\
  karypis@cs.umn.edu}

\maketitle


\begin{abstract}
  Grade prediction for future courses not yet taken by students is important as
  it can help them and their advisers during the process of course selection as
  well as for designing personalized degree plans and modifying them based on
  their performance. One of the successful approaches for accurately predicting
  a student's grades in future courses is Cumulative Knowledge-based Regression
  Models (CKRM). CKRM learns shallow linear models that predict a student's
  grades as the similarity between his/her knowledge state and the target
  course. However, prior courses taken by a student can have \black{different
    contributions when estimating a student's knowledge state and towards each
    target course, which} cannot be captured by linear models. Moreover, CKRM
  and other grade prediction methods ignore the effect of concurrently-taken
  courses on a student's performance in a target course. In this paper, we
  propose context-aware non-linear and neural attentive models that can
  potentially better estimate a student's knowledge state from his/her prior
  course information, as well as model the interactions between a target course
  and concurrent courses. Compared to the competing methods, our experiments on
  a large real-world dataset consisting of more than $1.5$M grades show the
  effectiveness of the proposed models in accurately predicting students'
  grades. Moreover, the attention weights learned by the neural attentive model
  can be helpful in better designing their degree plans.
\end{abstract}


\section{Introduction}

The average six-year graduation rate across four-year higher-education
institutions has been around 59\% over the past 15
years~\cite{kena2016condition,braxton2011understanding}, while less than half
of college graduates finish within four
years~\cite{braxton2011understanding}. These statistics pose challenges in
terms of workforce development, economic activity and national
productivity. This has resulted in a critical need for analyzing the available
data about past students in order to provide actionable insights to improve
college student graduation and retention rates.

One approach for improving graduation and retention rates is to help students
make good selections about the courses they register for in each term, such
that the knowledge they have acquired in the past would prepare them to succeed
in the next-term enrolled courses. Polyzou \etal~\cite{polyzou2016grade}
proposed course- and student-specific linear models that learns the importance
(or weight) or each previously-taken term towards accurately predicting the
grade in a future course. One limitation of this approach is that in order to
make accurate predictions, the model needs to have sufficient training data for
each (prior, target) tuple. Morsy \etal~\cite{morsy2017cumulative} developed
Cumulative Knowledge-based Regression Models (CKRM) that also build on the idea
of accumulating knowledge over time. CKRM predicts the student's grades as the
similarity between his/her knowledge state and the target course. Both the
student's knowledge state and the target course are represented as
low-dimensional embedding vectors and the similarity between them is modeled by
their inner product. The student's knowledge state is implicitly computed as a
linear combination of the so-called provided knowledge component vectors of the
previously-taken courses, weighted by his/her grades in them. Though CKRM was
shown to provide state-of-the-art grade prediction accuracy, it is limited in
that it learns shallow linear models that may not be able to capture the
\black{different contribution of each} prior course to estimate a student's
knowledge state, \black{as well as their different contributions towards each
  target course}. In addition, it does not consider the effect that
concurrently-taken courses can have on a student's performance in a target
course.

In this work, we develop context-aware non-linear and neural attentive models
that improve upon CKRM from two perspectives. First, they can model the
\black{different contribution of each} prior course to estimate the student's
knowledge state more accurately, by using two different approaches. In the
first approach, we hypothesize that each course provides a set of knowledge
components at a specific knowledge level. It uses a non-linear model that
aggregates the weighted prior course embeddings by employing a maximum-based
pooling layer along each component of the prior courses' embeddings. In the
second approach, we hypothesize that prior courses contribute differently
towards a target course, and that some of them may not be relevant to
it. Motivated by the success of neural attentive networks in different
fields~\cite{he2018nais,mei2018attentive,he2017neural,bahdanau2014neural,parikh2016decomposable,xiao2017attentional},
we learn attention weights for the prior courses that denote their importance
to a target course using two different activation
functions. 
Second, the proposed models consider the effect of the concurrently-taken
courses while predicting a student's grade in a target course. We hypothesize
that the knowledge provided by concurrent courses affect the knowledge required
by a target course. We model the interaction between the concurrent and target
course using non-linear and neural attentive models, as well.

The main contributions of this work are as follows:
\begin{enumerate}
\item We propose context-aware non-linear and neural attentive knowledge-based
  models for grade prediction that improve upon the linear CKRM model by: (i)
  using non-linear and neural attentive models to capture the \black{different
    contribution of each} prior course while aggregating their embeddings to
  compute a student's knowledge state, \black{as well as their different
    contributions towards each target course}; and (ii) modeling the effect of
  the concurrently-taken courses using non-linear and neural attentive
  models. To our knowledge, this is the first work to model the effect of the
  concurrently-taken courses in grade prediction.

\item We leverage the recent sparsemax activation function for the attention
  mechanism in the neural attentive models that produces sparse attention
  weights instead of soft attention weights.

\item We performed an extensive experimental evaluation on a real world dataset
  obtained from a large public university that spans a period of 16 years and
  consists of $\sim$1.5 grades. The results show that: (i) the proposed
  context-aware non-linear and neural attentive models outperform other
  baseline methods, including the previously-developed \ckrm\ method, with
  statistically significant improvements; (ii) the context-aware non-linear
  model outperforms the context-aware neural attentive model and all baselines
  in making less severe under-predictions; (iii) estimating a student's
  knowledge state via a non-linear or neural attentive model significantly
  outperforms estimating it via a linear model; (iv) learning sparse attention
  weights for the neural attentive model outperforms learning soft weights; (v)
  modeling the interactions between a target course and concurrent courses
  significantly improve the performance of the non-linear model and gives
  similar performance for the neural attentive model; and (vi) the neural
  attentive model was able to uncover the listed and hidden pre-requisite
  courses for target courses.
\end{enumerate}

\section{Definitions and Notations}

Boldface uppercase letters will be used to represent matrices (e.g., $\Gmat$) and
boldface lowercase letters to represent row vectors, (e.g., $\p$). The $i$th
row of matrix $\mathbf{P}$ is represented as $\p_i^T$, and its $j$th column is
represented as $\p_j$. The entry in the $i$th row and $j$th column of matrix
$\Gmat$ is denoted as $g_{i,j}$. A predicted value is denoted by having a hat over
it (e.g., $\hat{g}$).

Matrix $\Gmat$ will represent the $m \times n$ student-course grades matrix, where
$g_{s,c}$ denotes the grade that student $s$ obtained in course $c$, relative
to his/her average previous grade. Following the row-centering technique that
was first proposed by Polyzou \etal~\cite{polyzou2016grade}, we subtract each
student's grade from his/her average previous grade, since this was shown to
significantly improve the prediction accuracy of different models. As there can
be some students who achieved the same grades in all their prior courses, and
hence their relative grades will be zero, in this case, we assigned a small
value instead, i.e., 0.01. This is to prevent a prior course from not being
considered in the model computation. A student $s$ enrolls in sets of courses
in consecutive terms, numbered relative to $s$ from $1$ to the number of terms
in he/she has enrolled in the dataset. A set $\term_{s,w}$ will denote the set
of courses taken by student $s$ in term $w$.

\section{Related Work}

In this section, we review and identify several research areas that are highly
relevant to our work.

\subsection{Grade Prediction Methods}
\label{sec:related:grade-prediction}

Grade prediction approaches for courses not yet taken by students have been
extensively explored in the
literature~\cite{ren2017grade,ren2018ale,hu2018course,sweeney2016next,polyzou2016grade,morsy2017cumulative,elbadrawy2016domain}.
Sweeney \etal~\cite{sweeney2016next} investigated the use of recommender
systems techniques for grade prediction. They employed different methods, such
as matrix factorization, random forests and linear regression. Elbadrway
\etal~\cite{elbadrawy2016domain} developed several grade prediction and course
recommendation methods that make use of the student- and course-based academic
grouping information. Students can be grouped based on the colleges they
attend, their declared majors and/or their academic levels. Courses can be
grouped based on their subjects and/or levels, e.g. CSCI 5481 belongs to the
Computer Science subject and level 5. The authors hypothesized that grouping
students and/or courses into one of these finer-grained groups and
incorporating this information into matrix factorization, user-based
collaborative filtering, and popularity-based ranking, give more accurate grade
prediction and recommendation. To this end, the authors introduced the use of
local student and course biases into the aforementioned methods for both grade
prediction and course recommendation. Using the finer-grained grouping improved
the recommendation accuracy, but did not add much to the grade prediction
accuracy.

\subsubsection{Course-Specific Regression Models (CSR)}
\label{sec:related:csr}

A more recent and natural way to model the grade prediction problem is to model
the way the academic degree programs are structured. Each degree program would
require students to take courses in a specific sequencing such that the
knowledge acquired in previous courses are required for a student to perform
well in future courses. Polyzou \etal~\cite{polyzou2016grade} developed
course-specific linear regression models (CSR) that build on this idea. A
student's grade in a course is estimated as a linear combination of his/her
grades in previously-taken courses, with different weights learned for each
(prior, target) course pair. For a student $s$ and a target course $j$, the
predicted grade is estimated as:
\begin{equation}
  \label{eq:csr}
  \hat{g}_{s,j} = b_j + \sum_{i \in \mathcal{P}} w_{i, j} ~ g_{s, i},
\end{equation}
\noindent where $b_j$ is the bias term for course $j$, $w_{i, j}$ is the
weight of course $i$ towards predicting the grade of course $j$, $g_{s, i}$ is
the grade of student $s$ in course $i$, and $\mathcal{P}$ is the set of courses
taken by $s$ prior to taking course $j$. To achieve high prediction accuracy,
CSR requires sufficient training data for each (prior, target) pair, which can
hinder these models from good generalization.

\subsubsection{Cumulative Knowledge-based Regression Models (CKRM)}
\label{sec:related:ckrm}

Morsy \etal~\cite{morsy2017cumulative} developed Cumulative Knowledge-based
Regression Models (CKRM), which is also based on the fact that a student's
performance in a future course is based on his/her performance in the
previously-taken courses. It assumes that a space of knowledge components
exists such that each course provides a subset of these components as well as
requires the knowledge of some of these components from the student in order to
perform well in it. A student by taking a course thus acquires its knowledge
components in a way that depends on his/her grade in that course. The overall
knowledge acquired by a student after taking a set of courses is then
represented by a knowledge state vector that is computed as the sum of the
knowledge component vectors of those courses, weighted by his/her grades in
them. Let $\p_i$ denote the provided knowledge component vector for course
$i$. The knowledge state vector for student $s$ at term $t$ can be expressed as
follows:
\begin{equation}
  \label{eq:ks}
  \ks_{s,t} = \sum_{w = 1}^{t-1}  \xi(s, w, t) \sum_{i \in \term_{s, w}} \Big( g_{s,i} ~ \p_i \Big),
\end{equation}
\noindent where $g_{s,i}$ is the grade that student $s$ obtained on course
$i$, and $\xi(s, w, t)$ is a time-based exponential decaying function
designed to de-emphasize courses that were taken a long time ago.

Given a student's knowledge state vector prior to taking a course and that
course's required knowledge component vector, denoted as $r_j$, CKRM estimates
that student's expected grade in that course as the inner product of these two
vectors, i.e.,
\begin{align}
  \label{eq:ckrm}
  \hat{g}_{s,j} = b_j + \ks_{s,t}^T ~ \req_j,
\end{align}
\noindent where $b_j$ is as defined in Eq.~\ref{eq:csr}, and $\ks_{s,t}$ is
the corresponding knowledge state vector. These course-specific linear models
are estimated from the historical grade data and can be considered as capturing
and weighting the knowledge components that a student needs to have accumulated
in order to perform well in a course.


\subsection{Neural Attentive Models}

Neural networks have been used extensively in many fields, including, but not
limited to: Natural Language
Processing~\cite{parikh2016decomposable,bahdanau2014neural} and recommender
systems~\cite{xiao2017attentional,he2017neural,mei2018attentive,he2018nais}. The
attention mechanism has been recently introduced to neural network modeling and
was shown to improve the performance of different models. Instead of
aggregating the input object embeddings via a summation or mean pooling
function, which assumes equal contribution of all objects, the idea is to allow
the selected objects to contribute differently when compressing them to a
single representation. Neural attentive networks have been successfully applied
in many recommendation system techniques, such as factorization
machines~\cite{he2017neural,xiao2017attentional}, item-based collaborative
filtering~\cite{he2018nais}, and user-based collaborative
filtering~\cite{chen2017attentive}.

Part of our work relies on the attention mechanism, and leverages several
advances in this area. The most commonly-used activation function for the
attention mechanism is the softmax function, which is easily differentiable and
gives soft posterior probabilities that normalize to 1. A major disadvantage of
the softmax function is that it assumes that each object contributes to the
compressed representation, which may not always hold in some domains. To solve
this, we need to output sparse posterior probabilities and assign zero to the
irrelevant objects. Martins \etal~\cite{martins2016softmax} proposed the
sparsemax activation function, which has the benefit of assigning zero
probabilities to some output variables that may not be relevant for making a
decision. This is done by defining a threshold, below which small probability
values are truncated to zero. We also leverage the controllable sparsemax
activation function recently proposed by Laha \etal~\cite{laha2018controllable}
that controls the desired degree of sparsity in the output probabilities. This
is done by adding an L2 regularization term that is to be maximized in the loss
function. This will potentially encourage larger probability values for some
objects, moving the rest to zero.

\section{Non-linear and Neural Attentive Knowledge-based Models}
\label{mak-nak}

\ckrm~\cite{morsy2017cumulative} uses shallow linear models to aggregate the
prior courses' embeddings taken by a student in order to estimate his/her
knowledge state. \black{\ckrm\ assumes that each prior course contributes equally
  towards estimating a student's knowledge state}. We hypothesize that
\black{prior courses have different contributions when estimating a student's
  knowledge state, and they can also contribute differently towards a target
  course}. We develop two different approaches that can model these unique
contributions: a non-linear maximum knowledge-based model (Section
\ref{mak-nak:mak}), and a neural attentive knowledge-based model (Section
\ref{mak-nak:nak}).

\subsection{Maximum Knowledge-based Models}
\label{mak-nak:mak}

In this section, we develop a {\bf MA}ximum {\bf K}nowledge-based model (\mak),
which estimates a student's knowledge state by applying a maximum-based pooling
layer on the prior courses. We use \ckrm\ as the underlying model
(see Section~\ref{sec:related:ckrm}).

\subsubsection{Motivation}

Undergraduate degree programs are structured in a way such that earlier courses
provide basic knowledge that is built upon in the later courses that provide
more advanced knowledge. Consider 
two courses offered by a Computer Science department: Introduction to
Programming in C/C++ and Advanced Programming Principles. We would expect that
the introduction to programming course provides basic knowledge to programming
to freshman students who may be exposed to programming for the first time. The
advanced programming course builds on the knowledge acquired by the
introductory course, and provides more advanced knowledge components related to
programming principles and programming languages. When a student takes the
introductory then the advanced course, he/she can only acquire the maximum
knowledge components provided by both of them, since each course provides very
similar knowledge components, but at a different knowledge level.


\subsubsection{Maximum-based Pooling Layer for Prior Courses}

Based on our hypothesis explained above, we can estimate a student $s$'s
knowledge state at the beginning of term $t$ as follows:
\begin{equation}
  \label{eq:ks-max}
  \ks_{s, t} = \begin{bmatrix}
      \underset{i}{\max} \Big( \xi(s, w_{s,i}, t) ~ g_{s,i}  ~ p_{i, 1} \Big) \\
      . \\
      . \\
      . \\
      \underset{i}{\max} \Big( \xi(s, w_{s,i}, t) ~ g_{s,i} ~ p_{i, d} \Big)
    \end{bmatrix}, \forall i \in \term_{s,y} ~\textrm{for}~ y = 1, \dots, t-1,
\end{equation}
\noindent where $w_{s,i}$ is the relative term number when $s$ took course $i$,
$\xi(s, w_{s,i}, t)$ is a time-based exponential decaying function, $p_{i, z}$
is the $z$th entry in $\p_i$, $\term_{s,y}$ is the set of courses taken by $s$
in term $y$, and $d$ is the embedding size of the vector $\p$.

\subsection{Neural Attentive Knowledge-based Models}
\label{mak-nak:nak}

In this section, we develop a {\bf N}eural {\bf A}ttentive {\bf
  K}nowledge-based model (\nak), which applies an attention mechanism on prior
courses to learn individual weights for them that represent their importance to
a target course before aggregating them to estimate a student's knowledge
state. We also use \ckrm\ as the underlying model (see
Section~\ref{sec:related:ckrm}).

\begin{table*}[t]
  \renewcommand{\arraystretch}{1.3}
  \caption{Sample of prior and target courses for a Computer Science student at
    the University of Minnesota.}
  \begin{center}
    \begin{scriptsize}
  \begin{tabular}{lc}
    \toprule
    \multicolumn{1}{c}{Prior Courses} & Target Course \\
    \midrule
    \multirow{3}{*}{\parbox{11cm}{Calculus I, Beginning German, Operating
    Systems, Intermediate German I, University Writing, Introductory
    Physics, Peotics in Film,
    Program Design \& Development, Philosophy, Linear Algebra,
                      Internet Programming, Stone Tools to Steam Engines, Advanced Programming
                      Principles, Computer Networks}} & Intermediate German II
  \\
    \cmidrule{2-2}
                                        & Probability \& Statistics \\
    \cmidrule{2-2}
                  & Algorithms \& Data Structures \\
    \bottomrule
  \end{tabular}
  \label{tbl:sample-courses}
  \end{scriptsize}
\end{center}

\end{table*}

\subsubsection{Motivation}
\label{nak:motivation}

Consider a sample student who is declared in a Computer Science major and is in
his/her second or third year in college. Table~\ref{tbl:sample-courses} shows
the set of prior courses that this student has already take and the set of
courses that this student is planning on taking the next term. With CKRM
(Section~\ref{sec:related:ckrm}), all these prior courses would contribute
equally to predicting the grade of each target course. However, we can see
that, intuitively, from the courses' names, there are courses that are strongly
related to each target course and other courses that are irrelevant to it. For
instance, it is reasonable to expect that the Intermediate German II course is
more related to the Intermediate German I course than any of the other courses
that the student has already taken. Along the same lines, we expect that the
Algorithms and Data Structures course is more related to other Computer Science
courses, such as the Advanced Programming Principles and the Program Design and
Development courses. Assuming equal contribution among these prior courses can
hinder the grade prediction model from accurately learning the course
representations, and hence lead to poor predictions.

\subsubsection{Attention-based Pooling Layer for Prior Courses}
\label{nak:attention-prior}

In order to learn the different contributions of prior courses in estimating a
student's grade in a future course, we can employ the CSR technique (see
Section~\ref{sec:related:csr}) that learns the importance of each prior course
in estimating the grade of each future course. Thus, we would estimate a
knowledge state vector for each target course $j$, using the following
equation:
\begin{equation}
  \label{eq:nak}
  \ks_{s,t,j} = \sum_{w = 1}^{t-1} \sum_{i \in \term_{s,w}} \Big( a_{i, j}^p ~ g_{s,i} ~ \p_{i} \Big),
\end{equation}
\noindent where $a_{i, j}^p$ is a learnable parameter that denotes the
attention weight of course $i$ in contributing to student $s$'s knowledge state
when predicting his/her grade in course $j$. However, this solution requires
sufficient training data for each $(i, j)$ pair in order to be considered an
accurate estimation.

In order to be able to have accurate attention weights between all pairs of
prior and target courses, even the ones that do not appear together in the
training data, we propose to use the attention mechanism that was recently used
in neural networks~\cite{bahdanau2014neural,vaswani2017attention}. The main
idea is to estimate the attention weight $a_{i, j}^p$ from the embedding vectors
for courses $i$ and $j$.

In order to compute the similarity between the embeddings of prior course $i$
and target course $j$, we use a single-layer perceptron as follows:
\begin{equation}
  \label{eq:slp-prior}
  z_{i, j}^p = {\mathbf{h}^p}^T \textrm{RELU}(\mathbf{W}^p (\q_i \odot \req_j) + \mathbf{b}^p),
\end{equation}
\noindent where $\q_i = g_{s,i} \p_i$ denotes the embedding of the prior course
$i$, weighted by $s$'s grade in it, and
$\mathbf{W}^p \in \mathcal{R}^{l \times d}$ and
$\mathbf{b}^p \in \mathcal{R}^{l}$ denote the weight matrix and bias vector
that project the input into a hidden layer, respectively, and
$\mathbf{h}^p \in \mathcal{R}^{l}$ is a vector that projects the hidden layer
into an output attention weight, where $d$ and $l$ denote the number of
dimensions of the embedding vectors and attention network, respectively. RELU
denotes the Rectified Linear Unit activation function that is usually used in
neural attentive networks.

After computing the affinity vector $\mathbf{z}^p$ that represents the
similarity between each prior course and the target course, an activation
function is used to convert $\mathbf{z}^p$ to attention weights that follow a
probability distribution that sum up to 1. In the remaining of this section, we
explain the two activation functions that we used: the softmax and sparsemax
activation functions.

\paragraph{Softmax Activation Function}
\label{sec:methods:attention-prior:softmax}

The most common activation function used for computing these attention weights
is the softmax function~\cite{vaswani2017attention}. Given a vector of real
weights $\mathbf{z}$, the softmax activation function converts it to a
probability distribution, which is computed component-wise as follows:
\begin{equation}
  \label{eq:softmax-act}
  \textrm{softmax}_i(\mathbf{z}) = \frac{\exp(z_i)}{\sum_j \exp(z_j)}.
\end{equation}
\noindent We will refer to this method as {\bf NAK(soft)}.

\paragraph{Sparsemax Activation Function}
\label{sec:methods:attention-prior:sparsemax}

Although the softmax activation function has been used to design attention
mechanisms in many
domains~\cite{parikh2016decomposable,bahdanau2014neural,xiao2017attentional,he2017neural,mei2018attentive,he2018nais},
we believe that using it for grade prediction can degrade the accuracy of
prediction. Since a student enrolls in several courses, and each course
requires knowledge from one or a few other courses, we hypothesize that some of
the prior courses should have no effect, i.e., zero attention, towards
predicting a target course's grade. We thus leverage a recent advance, the
sparsemax activation function~\cite{martins2016softmax}, to learn sparse
attention weights. The idea is to define a threshold, below which small
probability values are truncated to zero. Let
$\triangle^{K-1} := \{ \mathbf{x} \in \mathbb{R}^K | \mathbf{1}^T\mathbf{x} =
1, \mathbf{x} \ge \mathbf{0}\}$ be the $(K-1)$-dimensional simplex. The
sparsemax activation function tries to solve the following equation:
\begin{equation}
  \textrm{sparsemax}(\mathbf{z}) = \underset{\mathbf{x} \in
    \triangle^{K-1}}{\textrm{argmin}} ~ \| \mathbf{x} - \mathbf{z} \|^2,
\end{equation}
\noindent which, in other words, returns the Euclidean projection of the input
vector $\mathbf{z}$ onto the probability simplex. We will refer to
this method as {\bf NAK(sparse)}.

In order to obtain different degrees of sparsity in the attention weights, Laha
\etal~\cite{laha2018controllable} developed a generic probability mapping
function for the sparsemax activation function, which they called
{\bf sparsegen}, and is computed as follows:
\begin{equation}
  \textrm{sparsegen}(\mathbf{z}; \gamma) = \textrm{argmin} ~ \| \mathbf{x} -
  \mathbf{z} \|^2 - \gamma \| \mathbf{x} \|^2,
\end{equation}
\noindent where $\gamma < 1$ controls the L2 regularization strength of
$\mathbf{x}$. An equivalent formulation for sparsegen was formed as:
\begin{equation}
  \label{eq:sparsegen-act}
  \textrm{sparsegen}(\mathbf{z}; \gamma) = \textrm{sparsemax} \big( \frac{\mathbf{z}}{1-\gamma} \big),
\end{equation}
\noindent which, in other words, applies a temperature parameter to the original
sparsemax function. Varying this temperature parameter can change the degree of
sparsity in the output variables. By setting $\gamma = 0$, sparsegen becomes
equivalent to sparsemax.

\section{Context-aware Non-linear and Neural Attentive Knowledge-based Models}

Another limitation of existing grade prediction methods is that they ignore the
effect of concurrently-taken courses. We hypothesize that the concurrent
courses can affect a student's grade in a target course. For instance, the
knowledge provided by concurrent courses can help a student in better
understanding the material given in a target course. In addition, since a
student's time is limited, the effort that he/she spends on a target course is
affected by the difficulty of courses taken concurrently with it. These
interactions create synergy and/or competition among a target course and
concurrently-taken courses We thus estimate a context-aware embedding for a
target course that we would like to predict a student's grade in, given the
courses taken concurrently with it. We utilize the proposed \mak\ and \nak\
models (Section~\ref{mak-nak}) as our underlying models.

To model the interactions between a target course and other courses taken
concurrently with it, we estimate a context-aware embedding for that target
course as follows:
\begin{equation}
  \label{eq:context-aware-req}
  \e_{j, w} = \x_{j,w} \odot \req_j,
\end{equation}
\noindent where $\x_{j, w}$ denotes the aggregated embedding of the courses
that are taken concurrently with $j$ in term $w$, $\odot$ denotes the Hadamard
product, and $\req_j$ denotes the required knowledge component vector for
target course $j$. To aggregate the concurrent courses' embeddings, we use
non-linear and neural attentive models similar to the ones developed
in Sections~\ref{mak-nak:mak} and~\ref{mak-nak:nak}, respectively.

\subsection{Context-aware Maximum Knowledge-based Models}
\label{methods:concurrent:cmak}

In this section, we develop a {\bf C}ontext-aware {\bf MA}ximum {\bf
  K}nowledge-based model (\cmak), which models the interactions between a
target and concurrent courses using \mak\ (Section~\ref{mak-nak:nak}) as the
underlying model.

The aggregated embedding of the courses that are taken concurrently with $j$ in
term $w$ is estimated by applying a maximum-based pooling layer on them,
similar to how we aggregated the prior courses' embeddings for the \mak\ model
(Section~\ref{mak-nak:mak}), and is computed as:
\begin{equation}
  \label{eq:cmak-x}
  \x_{j, t} = \begin{bmatrix}
    \underset{i}{\max} ~ p_{i, 1} \\
    . \\
    . \\
    . \\
    \underset{i}{\max} ~ p_{i, d}
  \end{bmatrix}, \forall i \in \term_{\{s, t\} \setminus \{j\}},
\end{equation}
\noindent where: $\p_{i, l}$ denotes the $l$th entry in the $\p_i$ vector,
where $\p_i$ denotes the embedding for concurrent course $i$. Note that we use
the same embedding vector $\p_i$ for representing both a prior and a concurrent
course.

\subsection{Context-aware Neural Attentive Knowledge-based Models}
\label{methods:concurrent:cnak}

In this section, we develop a {\bf C}ontext-aware {\bf N}eural {\bf A}ttentive
{\bf K}nowledge-based model (\cnak), which models the interactions between a
target and concurrent courses using \nak\ (Section~\ref{mak-nak:nak}) as the
underlying model.

To aggregate the concurrent courses' embeddings, we employ an attention
mechanism on them to learn the different contributions of each of them towards
the target course, similar to how we aggregated the prior courses' embeddings
for the \nak\ model (Section~\ref{nak:attention-prior}). The aggregated
embedding of the courses that are taken concurrently with $j$ in term $w$ is
computed as:
\begin{equation}
  \label{eq:cnak-x}
  \x_{j, w} = \sum_{i \in \term_{\{s, w\} \setminus \{j\}}} a_{i, j}^x \p_i,
\end{equation}
\noindent where $a_{j, t}^x$ is the attention weight for the concurrent course
$j$, and can be computed using the softmax (Eq.~\ref{eq:softmax-act}) or
sparsegen (Eq.~\ref{eq:sparsegen-act}) activation function.  The affinity
between concurrent course $i$ and target course $j$ is computed in a similar
way as in Eq.~\ref{eq:slp-prior}, i.e.,
\begin{equation}
  \label{eq:slp-concur}
  z_{i, j}^x = {\mathbf{h}^x}^T \textrm{RELU}(\mathbf{W}^x (\p_i \odot \req_j) + \mathbf{b}^x),
\end{equation}
\noindent where $\mathbf{W}^x \in \mathcal{R}^{l \times d}$,
$\mathbf{b}^x \in \mathcal{R}^{l}$ and $\mathbf{h}^x \in \mathcal{R}^{l}$
denote the attention network parameters for the concurrent courses, similar to
the ones defined in Eq.~\ref{eq:slp-prior}.

\section{Grade Prediction}

Given a student $s$'s representation at the beginning of term $t$ and a target
course $j$'s representation that he/she is interested in taking, we can
estimate $s$'s grade in $j$ for the different proposed methods in a similar way
to \ckrm\ as follows:
\begin{itemize}
\item Using \mak:
  \begin{equation}
    \label{eq:pred:mak}
    \hat{g}_{s,j} = b_j + \ks_{s,t}^T ~ \req_j,
  \end{equation}
  \noindent where: $\ks_{s, t}$ is as defined in Eq.~\ref{eq:ks-max} and $b_j$
  and $\req_j$ is as defined in Eq.~\ref{eq:ckrm}.

\item Using \nak:
  \begin{equation}
    \hat{g}_{s,j} = b_j + \ks_{s,t,j}^T ~ \req_j,
  \end{equation}
  \noindent where: $\ks_{s,t,j}$ is as defined in Eq.~\ref{eq:nak} and $b_j$ and
  $\req_j$ are as defined in Eq.~\ref{eq:pred:mak}.

\item Using \cmak:
  \begin{equation}
    \hat{g}_{s,j} = b_c + \ks_{s,t}^T ~ \e_{j, t},
  \end{equation}
  \noindent where: $b_j$ is as defined in Eq.~\ref{eq:pred:mak},
  $\ks_{s,t}$ is as defined in Eq.~\ref{eq:ks-max}, and $\e_{j, t}$ is
  as defined in Eq.~\ref{eq:context-aware-req}.

\item Using \cnak:
  \begin{equation}
    \label{eq:cnak}
    \hat{g}_{s,j} = b_c + \ks_{s,t,j}^T ~ \e_{j, t},
  \end{equation}
  \noindent where: $b_j$ is as defined in Eq.~\ref{eq:pred:mak}, $\ks_{s,t, j}$
  is as defined in Eq.~\ref{eq:nak}, and $\e_{j, t}$ is as defined in
  Eq.~\ref{eq:context-aware-req}.

\end{itemize}

\section{Model Optimization}
\label{optimization}

We use the mean squared error (MSE) loss function to estimate the parameters of
all our proposed models. We minimize the following regularized MSE loss:
\begin{equation}
  \label{eq:opt}
  L = -\frac{1}{2N} \sum_{{s,c} \in \Gmat} {(g_{s,c} - \hat{g}_{s,c})}^2 + \lambda||\Theta||^2,
\end{equation}
where $N$ is the number of grades in $\Gmat$. The hyper-parameter $\lambda$
controls the strength of L2 regularization to prevent overfitting, and
$\Theta = \{\{\mathbf{b}\}, \{\p_i\}, \{\req_i\}\}$ denotes the learnable
parameters for the \mak\ and \cmak\ models, $\Theta = $ $\{\{\mathbf{b}\}$,
$\{\p_i\}$, $\{\req_i\}$, $\mathbf{W}^p$, $\mathbf{b}^p$, $\mathbf{h}^p\}$,
denotes the learnable parameters for the \nak\ model, and $\Theta = $
$\{\{\mathbf{b}\}$, $\{\p_i\}$, $\{\req_i\}$, $\mathbf{W}^p$, $\mathbf{b}^p$,
$\mathbf{h}^p$, $\mathbf{W}^x$, $\mathbf{b}^x$, $\mathbf{h}^x\}$ denotes the
learnable parameters for the \cnak\ model, where $\mathbf{W}^p$,
$\mathbf{b}^p$, and $\mathbf{h}^p$ denote the attention mechanism parameters
for the prior courses, and $\mathbf{W}^x$, $\mathbf{b}^x$, and $\mathbf{h}^x$
denote the attention mechanism parameters for the concurrent courses.

The optimization problem is solved using AdaGrad
algorithm~\cite{duchi2011adaptive}, which applies an adaptive learning rate for
each parameter. It randomly draws mini-batches of a given size from the
training data and updates the related model parameters. The source code for
the proposed methods is available at: \url{www.test.com}.\footnote{The link
  for the source code will be available upon publication.}

\section{Evaluation Methodology}

\subsection{Dataset}

The data used in our experiments was obtained from the University of Minnesota
(UMN), which includes 96 majors from 10 different colleges, and spans the years
$2002$ to $2017$. At UMN, the letter grading system used is A--F, which is
converted to the 4--0 scale using the standard letter grade to GPA
conversion. We row-centered the student's grades in each term around his/her
GPA achieved in previous terms, which was shown to significantly improve the
prediction performance in~\cite{polyzou2016grade}. We removed any grades that
were taken as pass/fail. The final dataset includes $54,269$ students, $5,824$
courses, and $1,561,145$ grades in total.

\subsection{Generating Training, Validation and Test Sets}

At UMN, there are three terms, Fall, Summer and Spring. We used the data from
$2002$ to Spring $2015$ (inclusive) as the training set, the data from Spring
$2016$ to Fall $2016$ (inclusive) as the validation set, and the data from
Summer $2016$ to Summer $2017$ (inclusive) as the test set. For a target course
taken by a student to be predicted, that student must have taken at least four
courses prior to the target course, in order to have sufficient data to compute
the student's knowledge state vector. We excluded any courses that do not
appear in the training set from the validation and test sets.

\subsection{Baseline Methods}

We compared the performance of the proposed methods against the following grade
prediction methods:
\begin{enumerate}

\item {\bf{Matrix Factorization (MF):}} This method predicts the grade for
  student $s$ in course $i$ as:
  \begin{equation}
    \label{eq:mf}
  \hat{g}_{s,i} = \mu + sb_s + cb_i + \mathbf{u}_s^T ~ \mathbf{v}_i,
\end{equation}
\noindent where $\mu$, $sb_s$ and $cb_i$ are the global, student and course
bias terms, respectively, and $\mathbf{u}_s$ and $\mathbf{v}_i$ are the student
and course latent vectors, respectively. We used the squared loss function with
L2 regularization to estimate this
model.

\item {\bf KRM(sum):} This is the \ckrm\ method described in
  Section~\ref{sec:related:ckrm}, and the underlying model for our proposed
  models.

\item {\bf KRM(avg):} This is similar to the KRM(sum) method, except that the
  prior courses' embeddings are aggregated with mean pooling instead of
  summation. It was shown in later studies, e.g.~\cite{ren2018ale}, that it
  performs better than KRM(sum).

\end{enumerate}

We implemented KRM(sum) and KRM(avg) with a neural network architecture and
optimization similar to that of the proposed methods.

\subsection{Model Selection}

We performed an extensive search on the parameters of the proposed and baseline
models to find the set of parameters that gives us the best performance for
each model.

For all proposed and competing models, the following parameters were used. The
number of latent dimensions for course embeddings was chosen from the set of
values: \{8, 16, 32\}. The L2 regularization parameter was chosen from the
values: \{1e-5, 1e-7, 1e-3\}. Finally, the learning rate was chosen from the
values: \{0.0007, 0.001, 0.003, 0.005, 0.007\}. For the proposed \nak\ and
\cnak\ models, the number of latent dimensions for the MLP attention mechanism
was selected in the range [1, 4].  For the sparsegen activation function in
\nak\ and \cnak, the L2 regularization parameter $\gamma$ was chosen from the
values: \{0.5, 0.9\}. For KRM(sum), KRM(avg), \mak\ and \cmak, the
time-decaying parameter $\lambda$ was chosen from the set of values: \{0, 0.3,
0.5, 0.7, 1.0\}.

The training set was used for estimating the models, whereas the validation set
was used to select the best performing parameters in terms of the overall MSE
of the validation set.

\subsection{Evaluation Methodology}

The grading system used by UMN uses a 12 letter grade system (i.e., A, A-, B+,
$\ldots$ F). We will refer to the difference between two successive letter
grades (e.g., B+ vs B) as a \emph{tick}. We converted the predicted grades into
their closest letter grades. We assessed the performance of the different
approaches based on the Root Mean Squared Error (RMSE) as well as how many
ticks away the predicted grade is from the actual grade, which is referred to
as {\em Percentage of Tick Accuracy}, or PTA. We computed the percentage of
grades predicted with no error (zero tick), within one tick, and within two
ticks, which will be referred to as PTA0, PTA1, and PTA2, respectively. In
general, the grades that are predicted with at most one or two ticks error are
sufficiently accurate for the task of course selection.

In addition, we also report the percentage of grades predicted with severe
errors. We report two metrics: (i) severe under-predictions; and (ii) severe
over-predictions. {\em Severe under-predictions} will refer to the percentage
of grades that are predicted with three or more tick errors lower than the
actual corresponding grades. A severe under-prediction for a student in a
target course can result in an opportunity loss for that student who might
falsely think that he/she is not well qualified for taking that course. {\em
  Severe over-predictions} will refer to the percentage of grades that are
predicted with three or more tick errors higher than the actual corresponding
grades. A severe over-prediction for a student in a course can motivate that
student to take that course, incorrectly believing that he/she is well-prepared
for taking it and will perform well in it. This might cause a decrease in the
student's GPA or having to repeat that course at a later time.

\section{Experimental Results}

We present the results of our experiments to answer the following questions:
\begin{itemize}
\item[{\bf RQ1.}] How do the proposed context-aware non-linear and neural
  attentive models compare against the competing methods?
\item[{\bf RQ2.}] What is the impact of estimating a student's knowledge state
  via a non-linear or neural attentive model?
\item[{\bf RQ3.}] What is the impact of modeling the effect of concurrent
  courses on a student's performance in a target course?
\item[{\bf RQ4.}] Are we able to derive any insights about the importance of
  different prior courses to target courses from the neural attentive, i.e.,
  \nak, model?
\end{itemize}

\subsection{Performance against Competing Methods}

\begin{table}[t]
  \caption{Comparison with baseline methods.}
  \begin{center}
    \begin{tabular}{lllll}
      \toprule
      \multicolumn{1}{c}{Model} & \multicolumn{1}{c}{RMSE
                                  ($\downarrow$)}
      & \multicolumn{1}{c}{PTA0 ($\uparrow$)} & \multicolumn{1}{c}{PTA1 ($\uparrow$)} &
                                                                                        \multicolumn{1}{c}{PTA2 ($\uparrow$)}
      \\
      \midrule
      MF &  0.724 & 25.7 & 58.6 & 79.5 \\
      KRM(sum) & 0.584 & 32.6 & 70.1 & 87.7 \\
      KRM(avg) & 0.584 & 34.9 & 70.6 & 87.7 \\
      \hline
      \cmak & \ul{0.548}\dag\ (6.2) & 35.1 (0.6)
                                              & \ul{73.4} (4.0) & \ul{89.8} (2.4) \\
      \cnak & 0.569\dag\ (2.6) & \ul{35.5}\dag\
                                 (1.7) & 72.0
                                         (2.0)
                                                                                      & 88.7 (1.1) \\

      \bottomrule
    \end{tabular}
    \begin{minipage}{0.9\textwidth}
      Underlined entries represent the best performance in each
      metric. \dag\ denotes statistical significance over the best baseline
      model, using the Student's $t$-test with a $p$-level $< 0.05$. Numbers in
      parentheses denote the percentage of improvement over the best baseline
      value in each metric.
    \end{minipage}
    \label{tbl:res:competing-models}
  \end{center}
\end{table}

\begin{table}[ht]
  \caption{Severe under- and over-predictions by baseline and proposed models.}
  \begin{center}
    \begin{tabular}{lll}
      \toprule
      \multicolumn{1}{c}{Model} & \multicolumn{1}{c}{Severe Under-}
      & \multicolumn{1}{c}{Severe Over-} \\
                                & \multicolumn{1}{c}{predictions
                                  ($\downarrow$)} &
                                                    \multicolumn{1}{c}{predictions ($\downarrow$)} \\
      \midrule
      KRM(sum) & 5.4 & 6.9 \\
      KRM(avg) & 5.6 & 6.7 \\
      \hline
      \cmak & \ul{3.9} (27.4\%) & \ul{6.3} (5.4\%) \\
      \cnak & 4.9 (9.1\%) & 6.4 (3.4\%) \\
      \bottomrule
    \end{tabular}
    \begin{minipage}{0.9\textwidth}
      Underlined entries represent the best performance in each
      metric. Numbers in parentheses denote the percentage of improvement over
      the best baseline value in each metric.
    \end{minipage}
    \label{tbl:res:over-under-predictions}
  \end{center}
\end{table}

Table~\ref{tbl:res:competing-models} shows the performance of the proposed
models against the competing models ({\bf RQ1}). Among the baseline methods,
both KRM(sum) and KRM(avg) outperforms MF. KRM(avg) outperforms KRM(sum) in
predicting grades within no and one tick errors. Among all competing and
proposed methods, the proposed \cmak\ and \cnak\ models outperform all baseline
methods, with statistically significant improvements in some metrics, namely
the RMSE and PTA0 metrics. \cmak\ and \cnak\ achieve 6.2\% and 2.6\% lower
(better) RMSE, and 2.4\% and 1.1\% more accurate predictions within two tick
errors, respectively, than the best performing baseline method. This shows the
effectiveness of the proposed context-aware non-linear and neural attentive
models in more accurately predicting the grades of students in their future
courses than all competing methods. Comparing \cmak\ with \cnak, we see that
\cmak\ outperforms \cnak, achieving 3.7\% lower RMSE, and 1.2\% more accurate
predictions within two tick errors.

Table~\ref{tbl:res:over-under-predictions} shows the percentage of severe
under- and over-predictions that were made by the different baseline and
proposed models, denoting the grades that were predicted with three or more
tick errors lower and higher than the actual grades, respectively. Severe
under-predictions can result in an opportunity loss for students, urging them
not to take these under-predicted courses in fear of lowering their
GPAs. Severe over-predictions can result in urging them to take these
over-predicted courses that they may not be well-prepared for and may lower
their GPAs. For the severe under-predictions, both \cmak\ and \cnak\ outperform
the KRM variants significantly, achieving 27\% and 9\% less severe
under-predictions. For the severe over-predictions, both \cmak\ and \cnak\ also
outperform the KRM variants, achieving 5\% and 3\% less severe
over-predictions. Comparing \cmak\ with \cnak, we see that \cmak\ outperforms
\cnak, achieving 20\% less severe under-predictions, and 2\% less severe
over-predictions.  Since the grades in the data are row-centered around the
students' average grades and a course bias term is learned for each course, it
is hard for all these models to prevent severe over-predictions from occurring.

\begin{table}[ht]
  \caption{Effect of estimating students' knowledge states via non-linear and
    neural attentive models.}
  \begin{center}
    \begin{tabular}{lllll}
      \toprule
      \multicolumn{1}{c}{Model} &
                                  \multicolumn{1}{c}{RMSE
                                  ($\downarrow$)}
      & \multicolumn{1}{c}{PTA0 ($\uparrow$)} & \multicolumn{1}{c}{PTA1 ($\uparrow$)} &
                                                                                        \multicolumn{1}{c}{PTA2 ($\uparrow$)}
      \\
      \midrule
      KRM(sum) & 0.584 & 32.6 & 70.1 & 87.7 \\
      KRM(avg) & 0.584 & 34.9 & 70.6 & 87.7 \\
      \hline
      \mak & \ul{0.571}\dag\ (2.2) & 34.7 (-0.6) &
                                                   \ul{72.1}
                                                   (2.1)
                                                                                      & \ul{88.8}\dag\ (1.3) \\
      \hline
      \nak(soft) & 0.589 (-0.9) & \ul{35.3} (1.1) &
                                                    71.8
                                                    (1.7)
                                                                                      & 88.0 (0.3) \\
      \nak(sparse) & 0.574\dag\ (1.7) & \ul{35.3}\dag\
                                        (1.1) &
                                                \ul{72.1}
                                                (2.1)
                                                                                      & 88.7\dag\ (1.1) \\
      \bottomrule
    \end{tabular}
    \begin{minipage}{0.9\textwidth}
      Underlined entries represent the best performance in each metric. \dag\
      denotes statistical significance over the best baseline model, using the
      Student's $t$-test with a $p$-level $< 0.5$. Numbers in parentheses
      denote the percentage of improvement over the best baseline value in each
      metric.
    \end{minipage}
    \label{tbl:res:mak-nak-vs-krm}
  \end{center}
\end{table}

\subsection{Effect of Estimating Student's Knowledge State via Non-linear and
  Neural Attentive Models}

Table~\ref{tbl:res:mak-nak-vs-krm} shows the prediction accuracy of the \mak\
and \nak\ models compared to that of the \ckrm\ model, in terms of the RMSE and
PTA metrics ({\bf RQ2}). Both the \mak\ and \nak\ models outperform the KRM
variants, with some statistically significant improvements, showing the
importance of using more powerful, non-linear models that can model the
different contributions of prior courses when estimating a student' knowledge
state and towards each target course. Using a maximum-based pooling layer
(\mak) outperforms using an attention-based pooling layer (\nak) in the overall
RMSE only, implying that the former makes less severe errors in predicting the
grades.

Comparing the \nak\ models with the softmax and sparsemax activation functions,
we can see that learning sparse attention weights outperforms learning soft
attention weights. This is expected, since not all prior courses are relevant
to a target course, as illustrated later in the qualitative analysis
in Section~\ref{res:qual}.

\subsection{Effect of Modeling Concurrent Courses}

Table~\ref{tbl:res:concur-effect} shows the prediction accuracy of the proposed
context-aware models vs the proposed context-unaware models ({\bf RQ3}), in
terms of the RMSE and PTA metrics. \cmak\ outperforms \mak\ significantly,
achieving 4\% lower RMSE, and 1.1\% more accurate predictions within two tick
errors. On the other hand, \cnak\ slightly outperforms \nak\ with 0.9\% lower
RMSE, and achieves the same percentage of accurate predictions within two tick
errors. This shows that modeling the interactions between a target course and
concurrent courses helps in improving the prediction accuracy for a student's
grade in that target course.

\begin{table}[t]
  \caption{Effect of modeling concurrent courses on students' performance in
    target courses.}
  \begin{center}
    \begin{tabular}{lllll}
      \toprule
      \multicolumn{1}{c}{Model} &
                                  \multicolumn{1}{c}{RMSE ($\downarrow$)}
      & \multicolumn{1}{c}{PTA0 ($\uparrow$)} & \multicolumn{1}{c}{PTA1 ($\uparrow$)} &
                                                                                        \multicolumn{1}{c}{PTA2 ($\uparrow$)}
      \\
      \midrule
      \mak  & 0.571 & 34.7 & 72.1 & 88.8 \\
      \cmak & \ul{0.548}\dag\ (4.0) & 35.1\dag\ (1.2) &
                                                        \ul{73.4}\dag\
                                                        (1.8)
                                                                                      & \ul{89.8} (1.1)  \\
      \hline \hline
      \nak(sparse) & 0.574 & 35.3 & 72.1 & 88.7 \\

      \cnak & 0.569\dag\ (0.9) & \ul{35.5} (0.6) &
                                                   72.0
                                                   (-0.1)
                                                                                      & 88.7 (0.0)  \\

      \bottomrule
    \end{tabular}
    \begin{minipage}{0.9\textwidth}
      Underlined entries represent the best performance in each metric. \dag\
      denotes statistical significance over the corresponding non-context-aware
      model, while using the Student's $t$-test with a $p$-level $< 0.5$.
        \end{minipage}
      \end{center}
      \label{tbl:res:concur-effect}
\end{table}

\subsection{Qualitative Analysis on the Prior Courses Attention Weights}
\label{res:qual}

In this section, we study the behavior of the attention mechanism on prior
courses in the \nak\ model ({\bf RQ4}). Recall the motivational example for the
Computer Science student, discussed in Section~\ref{nak:motivation}. This
student had a set of prior courses and three target courses that we would like
to predict his/her grades in (See Table~\ref{tbl:sample-courses}). Using
KRM(sum) or KRM(avg), all the prior courses would contribute equally to the
prediction of each target course. Using our proposed \nak(sparse) model, the
attention weights for the prior courses with each target course are shown in
Table~\ref{tbl:nak-attn-weights}\footnote{These results were obtained by
  learning \nak\ models to estimate the actual grades and not the row-centered
  grades. Also, we used $\q_i = \p_i$ in Eq.~\ref{eq:slp-prior}. This allowed
  us to get more interpretable results.}.

We can see that, using the sparsegen activation function, only a few prior
courses are selected with non-zero attention weights, which are the most
relevant to each target course.

For the Intermediate German II course, we can see that the student's grade in
it is most affected by two courses: the Intermediate German I course, and the
University Writing course. The Intermediate German I course is listed as a
pre-requisite course for the Intermediate German II course. Though the
University Writing course is not listed as a pre-requisite course, after
further analysis, we found out that the Intermediate German II course requires
process-writing essays and are considered part of the grading system. Though
the German courses are not part of the student's degree program, and are taken
by a small percentage of Computer Science students, our \nak\ model was able to
learn accurate attention weights for them.

The other two target courses, Probability and Statistics, and Algorithms and
Data Structures, have totally different prior courses with the largest
attention weights, which are more related to them.

These results illustrate that the proposed \nak\ model was able to uncover the
listed as well as the hidden/informal pre-requisite courses without any
supervision given to the model.

\begin{table}[ht]
  \renewcommand{\arraystretch}{1.3}
  \caption{The attention weights of the prior courses with each target course
    for the sample student from Table~\ref{tbl:sample-courses}.}
  \begin{center}
  \begin{tabular}{p{9.5cm}c}
    \toprule
    \multicolumn{1}{c}{Prior Courses} & Target Course \\
    \toprule
    Intermediate German I: \red{0.6980}, University Writing: \red{0.3020} & Intermediate German II \\
    \midrule
    Calculus I: \red{0.4737}, Physics: \red{0.3794}, Program Design \&
                                                                       Development: \red{0.0717},
                                                                       Operating
                                                                       Systems: \red{0.0497},
                                                                       Computer
                                                                       Networks: \red{0.0255} & Probability \&
                                                                                 Statistics
    \\
    \midrule
    Operating Systems: \red{0.2927},
    Advanced Programming Principles: \red{0.2582}, Linear Algebra:
    \red{0.2313}, Physics: \red{0.2178}  & Algorithms \& Data Structures \\
    \bottomrule

  \end{tabular}
  \begin{minipage}{0.9\textwidth}
    Prior courses are sorted in non-increasing order w.r.t. to their attention
    weights with each target courses for clarity purposes.
  \end{minipage}
  \label{tbl:nak-attn-weights}
\end{center}

\end{table}

\section{Conclusion and Discussion}

In this work, we presented context-aware non-linear and neural attentive models
that improve upon the previously developed \ckrm\ method, by: (i) using more
powerful, non-linear models that can model the different contributions of prior
courses when estimating a student's knowledge state and towards each target
course; and (ii) modeling the interactions between a target course and
concurrently-taken courses. The experiments showed that the proposed models
significantly outperformed all baseline methods. In addition, the proposed
neural attentive models are able to capture the listed as well as the hidden
pre-requisite courses for the target courses, which can be better used to
design better degree plans.

\black{While the \cmak\ and \cnak\ methods have significantly outperformed the
  existing methods, they, like the other methods, ignore some contextual
  information, such as the course's instructor, and the student's academic
  level. This information can further boost the accuracy of grade
  prediction. We plan to incorporate this information into our proposed models
  in the future.}

\section*{Acknowledgement}

This work was supported in part by NSF (1447788,
1704074, 1757916, 1834251), Army Research Office (W911NF1810344), Intel Corp,
and the Digital Technology Center at the University of Minnesota. Access to
research and computing facilities was provided by the Digital Technology Center
and the Minnesota Supercomputing Institute, \url{http://www.msi.umn.edu}.

\bibliographystyle{acmtrans}
\bibliography{refs}

\end{document}